# Joint-bone Fusion Graph Convolutional Network for Semi-supervised Skeleton Action Recognition

Zhigang Tu, *Member, IEEE,* Jiaxu Zhang, *Student Member, IEEE,* Hongyan Li, *Member, IEEE,* Yujin Chen, *Member, IEEE,* and Junsong Yuan, *Fellow, IEEE*

***Abstract*—In recent years, graph convolutional networks (GCNs) play an increasingly critical role in skeleton-based human action recognition. However, most GCN-based methods still have two main limitations: 1) They only consider the motion information of the joints or process the joints and bones separately, which are unable to fully explore the latent functional correlation between joints and bones for action recognition. 2) Most of these works are performed in the supervised learning way, which heavily relies on massive labeled training data. To address these issues, we propose a semi-supervised skeleton-based action recognition method which has been rarely exploited before. We design a novel correlation-driven joint-bone fusion graph convolutional network (CD-JBF-GCN) as an encoder and use a pose prediction head as a decoder to achieve semi-supervised learning. Specifically, the CD-JBF-GC can explore the motion transmission between the joint stream and the bone stream, so that promoting both streams to learn more discriminative feature representations. The pose prediction based auto-encoder in the self-supervised training stage allows the network to learn motion representation from unlabeled data, which is essential for action recognition. Extensive experiments on two popular datasets, i.e. NTU-RGB+D and Kinetics-Skeleton, demonstrate that our model achieves the state-of-the-art performance for semi-supervised skeleton-based action recognition and is also useful for fully-supervised methods.**

## I. Introduction

HUMAN action recognition has become one of the most important tasks in the computer vision field as it has a wide range of applications in intelligent video surveillance [1], [2], human-machine interaction [3], [4], medical service [5], etc. Skeleton-based human action recognition technique has attracted great attention due to its robustness against changes in camera viewpoints, human body scales, and interference of backgrounds. Besides, its recognition accuracy has been enhanced by exploiting approaches to effectively extract spatial-temporal features of skeleton sequences [6]–[15]. However, although increasingly human skeleton data are collected by depth camera and human pose estimation algorithms [16], [17], these unlabeled data cannot be directly used by existing skeleton-based human action recognition methods, because they heavily rely on the manual annotations which are costly to acquire. Especially when lacking of the appearance information, it is difficult to recognize the subtle movement of human skeleton and manually distinguish the skeleton action. Semi-supervised learning, a powerful approach that can explore useful information from unlabeled data, is widely used for recognizing human actions in RGB data [18]–[21], but it is rarely utilized for skeleton data analysis. To sum up, the skeleton-based action recognition with semi-supervised learning has two major challenges: one is how to extract discriminative features from skeleton sequences, and the other is how to make full use of unlabeled skeleton data. Our motivation is to address these issues.

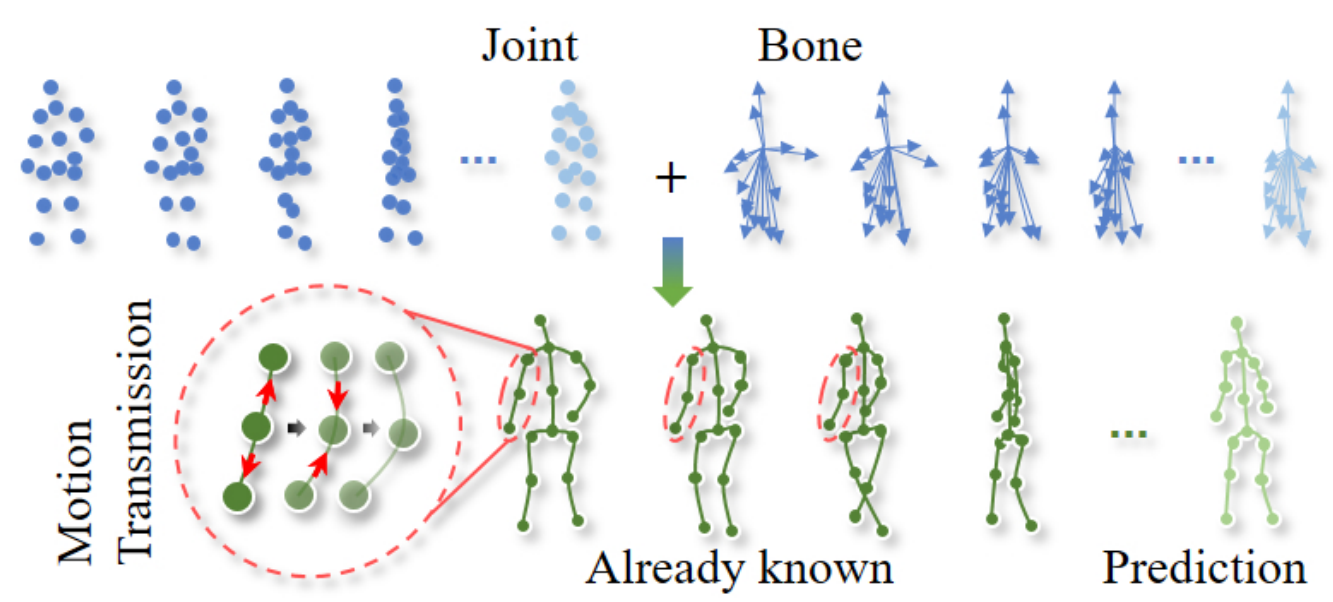

Fig. 1. Visualization of joint coordinates sequence, bone vectors sequence and motion transmission between joints and bones. “Joint+Bone” represents to connect joints and bones via their natural connection.

Naturally, the skeleton data represents the human action as a sequence of 2D or 3D coordinates of the main body joints, and these main joints are connected according to the physical structure of the human body, where this connection is called the “bone”. Therefore, the skeleton data expresses human actions via the motion information of joints and bones. Recently, GCN-based methods have become the mainstream to handle the problem of skeleton-based human action recognition [22]–[24], which can alternately perform spatial and temporal convolution operations to jointly extract the spatial-temporal information of the skeleton graph sequence. Based on the ST-GCN, many variants were explored [25]–[30], which typically introduce some incremental modules, e.g. the attention module [25], [31], the context-aware module [29], and the semantics-guided module [28], to enhance the network capacity. However, these ST-GCN based methods have two main drawbacks:

1) Most of these methods only consider the joint coordinates

Zhigang Tu, Jiaxu Zhang are with The State Key Laboratory of Information Engineering in Surveying, Mapping and Remote Sensing, Wuhan University, Wuhan 430079, China. (Email: zhigangtu@whu.edu.cn, zjiaxu@whu.edu.cn).

Hongyan Li is with the School of Information Engineering, Hubei University of Economics, Wuhan 430205, China.

Yujin Chen is with the Department of Informatics, Technical University of Munich, Garching 85748, Germany.

Junsong Yuan is with the Computer Science and Engineering department, State University of New York at Buffalo, USA. (Email: jsyuan@buffalo.edu).

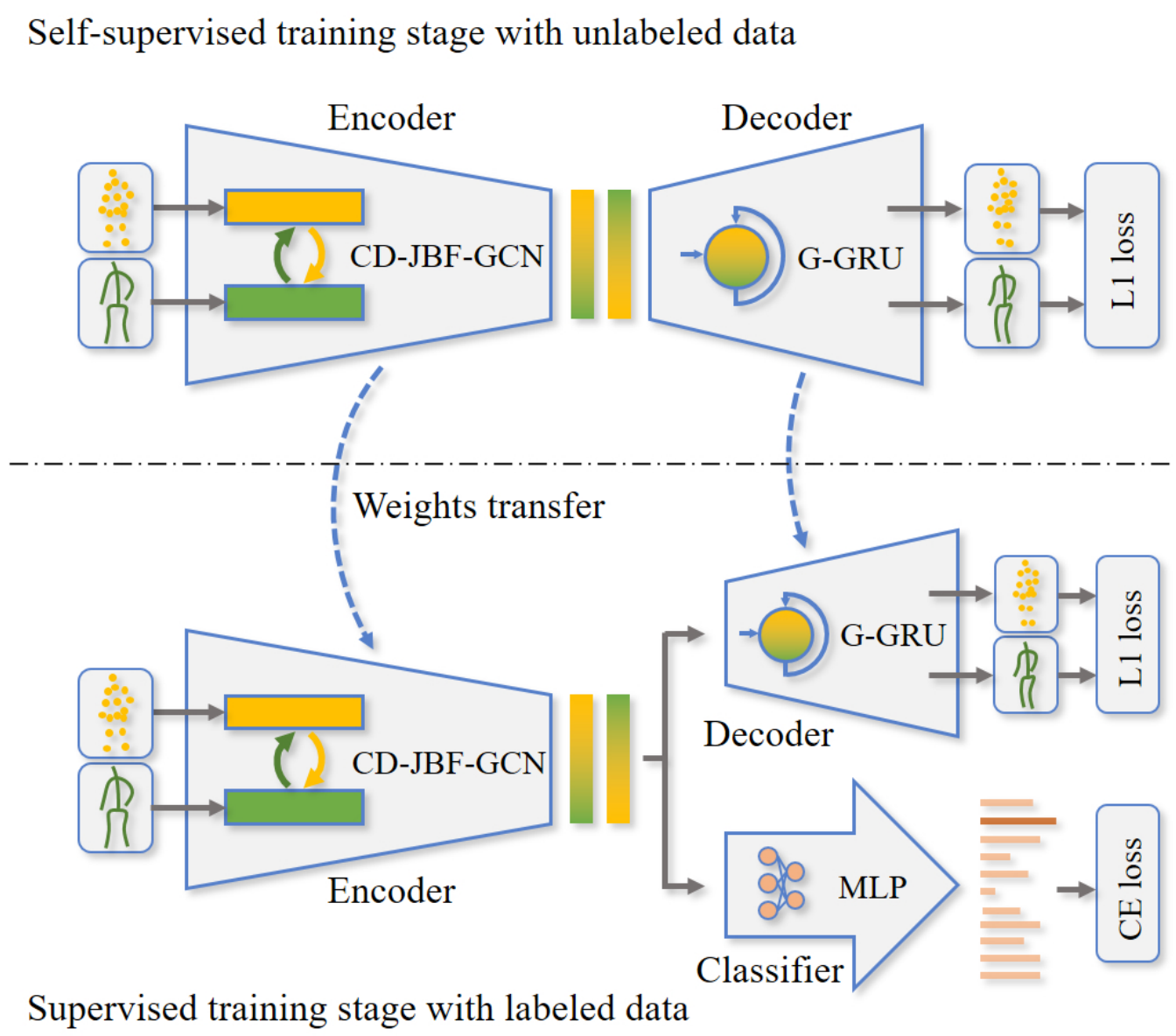

Fig. 2. Overview of our proposed CD-JBF-GCN and semi-supervised learning method for skeleton-based action recognition. In the training phase, we use both the annotated data and unannotated data to train the model. For the training data without action labels, we use a L1 loss to optimize the Encoder and Decoder. For the training data with action labels, we use a composite loss to optimize the Encoder, Decoder, and Classifier. In the testing phase, the Encoder and the Classifier are utilized to classify human actions that are represented by the skeleton sequence.

but ignore the bone vectors. Recently, 2s-AGCN [32] uses bone vectors as the input of the second stream, but the joint stream and the bone stream are completely independently treated, which limits its capability to fully explore the motion transmission of the skeleton structure. As can be seen in Fig. 1, if we treat the changes of the joint coordinates or bone vectors separately, it is difficult to distinguish human actions. It becomes easier for us to accurately capture the motion information of the skeleton and recognize human actions when combining joints and bones via their natural connection. Considering the potential functional correlation between the movement of joints and bones (see Eq. 4), leveraging the motion transmission of joints and bones may be beneficial for the action presentation and recognition.

2) Most of the above-mentioned approaches are supervised which heavily rely on massive labeled training data, but the labeled data are usually expensive to obtain. Hence, how to effectively learn motion representation from the easily accessible unlabeled skeleton data is challenging. As shown in Fig. 1, from a few skeleton frames of an action (e.g. walking), human can predict the action in future few frames. This is because the motion information (e.g. speed, acceleration) of joints and bones is continuous and regular, which is crucial for recognizing human actions. Existing researches [33], [34] have not yet used this phenomenon to design the self-supervised learning method.

Inspired by the recent work AS-GCN [26] which utilizes a pose prediction head to help capture more detailed motion patterns, we propose a novel correlation-driven joint-bone fusion graph convolutional network (CD-JBF-GCN) and use a pose prediction head in the self-supervised training stage, to perform semi-supervised learning for skeleton-based human action recognition. Specifically, similar to 2s-AGCN [32], we design an advanced two-stream network, which processes the joint and bone information separately but with the two streams interactively. This is achieved by building a bridge between the two streams according to the natural connection of joints and bones, where the bridge can promote both streams to learn more discriminative feature representations as it is able to transmit and fuse motion information between the two streams. We call this bridge is CD-JBF-GC. Moreover, we design an encoder-decoder structure to learn motion representation of the skeleton sequence in a semi-supervised way. This encoder-decoder structure can be self-supervised trained by minimizing the gap between the predictions and the original skeleton sequence with unlabeled samples, then the weights of the encoder can be transferred to supervised training with a few labeled samples. The proposed GCN-based network, i.e. CD-JBF-GCN, has two heads with different structures: one head is G-GRU based decoder, and the other is Multi-layer Perceptron (MLP) based classifier (see Fig. 2). Notably, unlike 2s-AGCN [32] which trains the model independently, the joint stream and bone stream in our encoder structure are trained simultaneously and interacts with each other.

In summary, the main contributions of our method are three-fold:

- A correlation-driven joint-bone fusion graph convolution (CD-JBF-GC) is proposed to simulate motion transmission between joints and bones, which is able to transmit and fuse information between the two streams;
- An auto-encoder structure is designed by introducing a G-GRU based pose prediction head, which can be self-supervised trained with unlabeled samples and effectively learn the motion representation of human actions;
- Experiments on two large-scale datasets demonstrated that the explored CD-JBF-GCN obtains the state-of-the-art performance for semi-supervised skeleton-based action recognition, and it is also beneficial for fully-supervised learning.

The remaining paper is organized as follows. The related work is presented in Section II. Section III introduces the backgrounds. Section IV illustrates the proposed CD-JBF-GCN and our semi-supervised method. Section V describes the experimental settings and analyzes of the results. The conclusion is given in Section VI.

## II. Related Work

### A. Skeleton-based Action Recognition

Conventional skeleton-based action recognition methods, which usually utilize handcrafted features [11], [35], or use RNN [3], [6], [30], [36]–[38] and CNN [8], [39]–[42] to learn deep features of the skeleton sequence, to help recognize human actions. Vemulapalli *et al.* [35] represented the relative 3D rotations between various body parts by designing rolling maps. Liu *et al.* [3] extended the RNN-based methods to spatial-temporal domains to analyze the action-related information. Zhu *et al.* [42] proposed a cuboid CNN to fully exploit the local movements of human joints in skeleton actions. However, these methods cannot effectively extract the spatial-temporal correlation of skeleton sequences and also cannot fully explore the graph structure of human body. Yan *et al*. [24] firstly presented a GCN-based method ST-GCN, which significantly boosts the performance of skeleton-based action recognition. Then later, many studies have been carried out to improve ST-GCN. For example, Li *et al*. [26] proposed an encoder-decoder structure to capture richer joint correlations. Cheng *et al*. [43] introduced a graph-based shift operation to provide flexible receptive fields and used the point-wise convolutions to lighten the computational complexity. Zhang *et al*. [31] explored a spatial attentive and temporal dilated GCN to extract the features of skeleton sequences with different spatial attention weights and temporal scales. Liu *et al*. [44] designed an unified graph convolutional operator for long-range spatial-temporal modeling. However, these methods only treat bones as the adjacency matrix of joints, while ignoring the direction and motion information that contains in the bone. Shi *et al*. [32] took bones as the second stream and constructed a two-stream network, but they treat the joint stream and the bone stream completely independent. Moreover, DGNN [45] and GECNN [46] were proposed to explore the relationship between joints and bones, but they did not discuss different ways to fuse features from the perspective of motion transmission. Following the GCN-based methods, our model connects joints and bones according to the physical structure of the human body, and utilizes the motion transmission function to extract the spatial-temporal motion information more effectively.

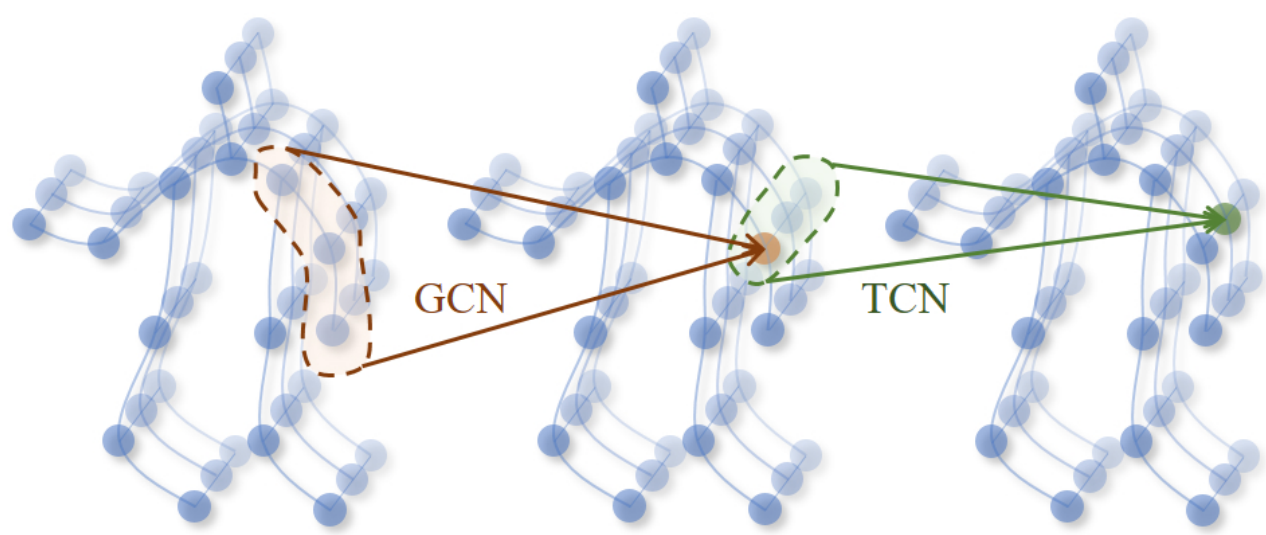

Fig. 3. Illustration of the process of an ST-GC layer. The GCN operate on the spatial dimension and the TCN operate on the temporal dimension.

### *B. Semi-supervised Action Recognition*

By regularizing the features or solving the pretext tasks, semi-supervised learning for action recognition aims to learn motion representation from the unlabeled data. Many studies have designed semi-supervised learning methods to process the unlabeled RGB data [18], [19], [47]–[49]. [21] and [20] proposed to learn the video representation by predicting motion flows. [1] designed a novel convolutional auto-encoder structure to separately capture the spatial and temporal information in raw videos. According to the fact that changing the frame rate of the video does not change the human actions, [49] constructed a two-stream contrastive model in temporal domain to make use of unlabeled videos at two different speeds. These methods perform well on the RGB video data but are not suitable for long-term skeleton sequences. Different from the RGB video, which contains rich scene information and appearance information, skeleton data focuses on the motion information of human body. Therefore, how to learn the movement patterns of joints and bones of the human body is very important for semi-supervised learning of skeleton actions. Zheng *et al*. [34] proposed a joint inpainting method to learn the features from the unlabeled skeleton data, and then Si *et al*. [33] followed this method. However, this self-supervised method that inpainting the key joints, cannot effectively learn the human body's motion information and is unfitted for GCN-based networks. Lin *et al*. [50] constructed a multi-task self-supervised learning architecture for skeleton data, but they can not effectively extract the joint-bone fused features with a reasonable network structure that aims at action recognition. In addition, [33], [34], [50] still use the RNN-based backbone, which is poorly to extract the spatial-temporal information. Even worse, there are few studies focus on the semi-supervised skeleton action recognition, because it is a challenging task to learn motion representation by using the unlabeled long-term skeleton sequences. In this work, we design a novel self-supervised framework to learn the feature representation of human actions. After fine-tuning with a small amount of labeled data, our semi-supervised model exceeds the state-of-the-art by a significant margin.

### *C. Graph Convolutional Network*

For the real world data, where many are in the irregular non-Euclidean space, e.g. the molecular structure [51], the transportation network [52], the social network [53], and the skeleton graph [24], making it is necessarily to improve the feature representation ability of the deep model in the non-Euclidean space. Accordingly, studying the neural network with the graph structure has gained increasingly research interest.

Generally, the neural networks operate on the graph following two streams: 1) the spectral domain, which is considered in the form of spectral analysis. *E.g.* Scarselli *et al.* firstly proposed a graph neural network (GNN) to handle the graph structure data [54] based on the conventional neural network. GNN can aggregate vertex information with the help of the manually designed rules in the spectral domain and can learn graph feature representations in a data-driven way. Defferrard *et al*. expanded the convolution operation into the non-Euclidean space by using the Fourier transform on the graph structure, and presented the graph convolutional network (GCN) for graph classification [55]. However, these spectral domain methods leading to an expensive computational cost. 2) The spatial domain, which directly operates on the graph vertexes. *E.g.* Niepert *et al*. [56] conducted the convolution operation on locally connected vertexes which approximates to the image-based methods. Monti *et al*. constructed an effective spatial domain GCN [57], which avoids complex spectral analysis steps, *e.g.* the Fourier transform and the Chebyshev polynomial approximation, to reduce the computational cost significantly. Following [24] and the spatial domain methods, our work also uses the GCN-based model to handle the skeleton data in non-Euclidean space directly.

## III. Background

### A. Problem Formulation

In this paper, we use $\mathcal{G} = (\mathcal{V}, \mathcal{E})$ to represent the skeleton graph, where $\mathcal{V}$ denotes the vertexes and $\mathcal{E}$ means the edges. For the joint graph, $\mathcal{V}$ is the set of $n$ joints and $\mathcal{E}$ is the set of $m$ bones. For the bone graph, on the contrary, $\mathcal{V}$ is the set of $m$ bones and $\mathcal{E}$ is the set of $n$ joints. We consider the adjacency matrix of the skeleton graph as $A \in \{0,1\}^{n\times n}$, where $A_{i,j} = 1$ if the *i*-th and the *j*-th vertexes are connected and 0 otherwise. Let $D \in \mathbb{R}^{n\times n}$ be the diagonal degree matrix, where $D_{i,i} = \sum_j A_{i,j}$. Following the work of ST-GCN [24], we divide one root vertex and its one-order neighbors into three sets to better express the structural information of the skeleton graph. In this way, $A$ is accordingly classified to be $A^r$, $A^c$ and $A^f$. We denote the partition group set as $P = \{r, c, f\}$ and $\sum_{p\in P} A^p = A$. Let $X_j \in \mathbb{R}^{n\times 3\times T}$ be the 3D joint positions and $X_b \in \mathbb{R}^{m\times 3\times T}$ be the 3D bone vectors across $T$ frames. Noticeably, the raw bone vectors are centered on the origin of the coordinates, which do not contain the structural information of the human body (see Fig. 1). In this work, we split the training data into two subsets: a labeled training set denoted as $L$ and an unlabeled training set denoted as $U$. Usually, $L$ is smaller than $U$.

### B. Spatial-temporal GCN

According to 2s-AGCN [32], in the spatial dimension, the convolution operation on the skeleton graph is:

$$X^{'} = \sum\nolimits_{p\in P} \left(\widetilde{A^p} + B^p + C^p\right) X W^p, \tag{1}$$

where $X \in \mathbb{R}^{n\times d_{in}}$ and $X^{'} \in \mathbb{R}^{n\times d_{out}}$ are the input and output features of all joints in one frame respectively, and $d_{in}$ and $d_{out}$ are their channel dimension. $\widetilde{A^p} = D^{p^{-\frac{1}{2}}} A^p D^{p^{-\frac{1}{2}}} \in \mathbb{R}^{n\times n}$ is the normalized adjacency matrix of each partition set. $W^p \in \mathbb{R}^{d_{in}\times d_{out}}$ is the trainable weights for each partition in the spatial GCN. $B^p \in \mathbb{R}^{n\times n}$ is an adaptive adjacency matrix that can be optimized in the training process just like other trainable parameters in the model. $C^p \in \mathbb{R}^{n\times n}$ is a feature-correlation-dependent adjacency matrix which can represent how strong the connection between two vertexes is. We calculate $C^p$ as follows:

$$C^p = softmax\left(\left(XW_\phi^p\right)\left(W_\theta^{p^T} X^T\right)\right), \tag{2}$$

where $W_\phi \in \mathbb{R}^{d_{in}\times n}$ and $W_\theta \in \mathbb{R}^{d_{in}\times n}$ are the trainable parameters of the embedding functions. The $softmax(\cdot)$ function operates on each row of the matrix.

In the temporal dimension, to capture the time sequential information, we operate a 2D convolution on the output feature map of the spatial convolution with a $K_t \times 1$ kernel (TCN in Fig. 3 and Fig. 4), $K_t$ denotes the kernel size of the temporal dimension. The process of a ST-GC layer is shown in Fig. 3.

### C. Graph Gated Recurrent Unit

As Li *et al.* [58] stated that the function of a Graph Gated Recurrent Unit (G-GRU) is to learn and update the hidden states of the graph vertexes under the guidance of the graph structure. Let $X^{(t)}$ be the initial state of G-GRU, and $H^{(t)}$ denotes the online skeleton motion feature. $G-GRU\left(X^{(t)}; H^{(t)}\right)$ works as follows:

$$\begin{aligned} r^{(t)} &= sigmoid\left(r_{in}\left(X^{(t)}\right) + r_{hid}\left(A_H H^{(t)} W_H\right)\right), \\ u^{(t)} &= sigmoid\left(u_{in}\left(X^{(t)}\right) + u_{hid}\left(A_H H^{(t)} W_H\right)\right), \\ c^{(t)} &= tanh\left(c_{in}\left(X^{(t)}\right) + r^{(t)} \odot c_{hid}\left(A_H H^{(t)} W_H\right)\right), \\ H^{(t+1)} &= u^{(t)} \odot H^{(t)} + \left(1 - u^{(t)}\right) \odot c^{(t)}, \end{aligned} \tag{3}$$

where $r_{in}(\cdot)$, $r_{hid}(\cdot)$, $u_{in}(\cdot)$, $u_{hid}(\cdot)$, $c_{in}(\cdot)$, and $c_{hid}(\cdot)$ are the trainable linear mappings. $W_H$ denotes the trainable weights.

## IV. Method

### A. Correlation-driven Joint-bone Fusion Graph Convolution

Existing models either without using the bone information [24], [26], [30], [59], or separate joints and bones for independent processing [25], [32], [60], [61]. These methods have not fully explored the movement transmission between joints and bones, nor make full use of the physical structure of the human body. Obviously, there is a hidden functional correlation between the movement of joints and the movement of bones. This function can be expressed as follows:

$$\begin{aligned} \Delta x_j^{'} &= f_j\left(\Delta x_j, x_b\right), \\ \Delta x_b^{'} &= f_b\left(\Delta x_b, x_j\right), \end{aligned} \tag{4}$$

where $\Delta x_j$ is the motion information of a joint $x_j$, $x_b$ is the bone vector connected to this joint, $\Delta x_j^{'}$ is the motion information of the joint adjacent to $x_j$. The movement relationship between $\Delta x_j$ and $\Delta x_j^{'}$ is transmitted through the bone $x_b$ using a certain function $f_j(\cdot)$. For bones, the principle is similar. The previous researches have not explored this hidden functional correlation, so they cannot reveal the internal information transmission between joints and bones. In this work, we still use the two-stream network for joints and bones, but we build a bridge between the two streams to simulate the transmission of motion information on joints and bones. This bridge is called CD-JBF-GC. Let $X_j \in \mathbb{R}^{n\times d}$ and $X_b \in \mathbb{R}^{m\times d}$ denote the feature maps respectively extracted from the joint stream and the bone stream. Taking the joint stream as an example, inspired by Eq. 4, we use the following method to update it:

$$X_j^{'} = \sum_{p\in P} f\left(A^p X_j, V^p X_b\right), \tag{5}$$

where $A^p = \widetilde{A^p} + B^p + C^p \in \mathbb{R}^{n\times n}$ is the adjacency matrix of joints. $V^p \in \mathbb{R}^{n\times m}$ is the correlation matrix between joints and bones which is also composed of three additive matrices, that is $V^p = \widetilde{V^p} + Y^p + Z^p$. $\widetilde{V^p}$ is the normalized $V^p$. $Y^p \in \mathbb{R}^{n\times m}$ is a trainable matrix (like $B^p$) and $Z^p \in \mathbb{R}^{n\times m}$ is a vertex-dependent matrix (like $C^p$). We use $A^p$ and $V^p$ for simplicity. $f(\cdot)$ is the feature transmission function. The flow chart of the CD-JBF-GC module is shown in Fig. 4. Next, we will introduce how to construct $V^p$, and discuss the implementation and performance of $f(\cdot)$.

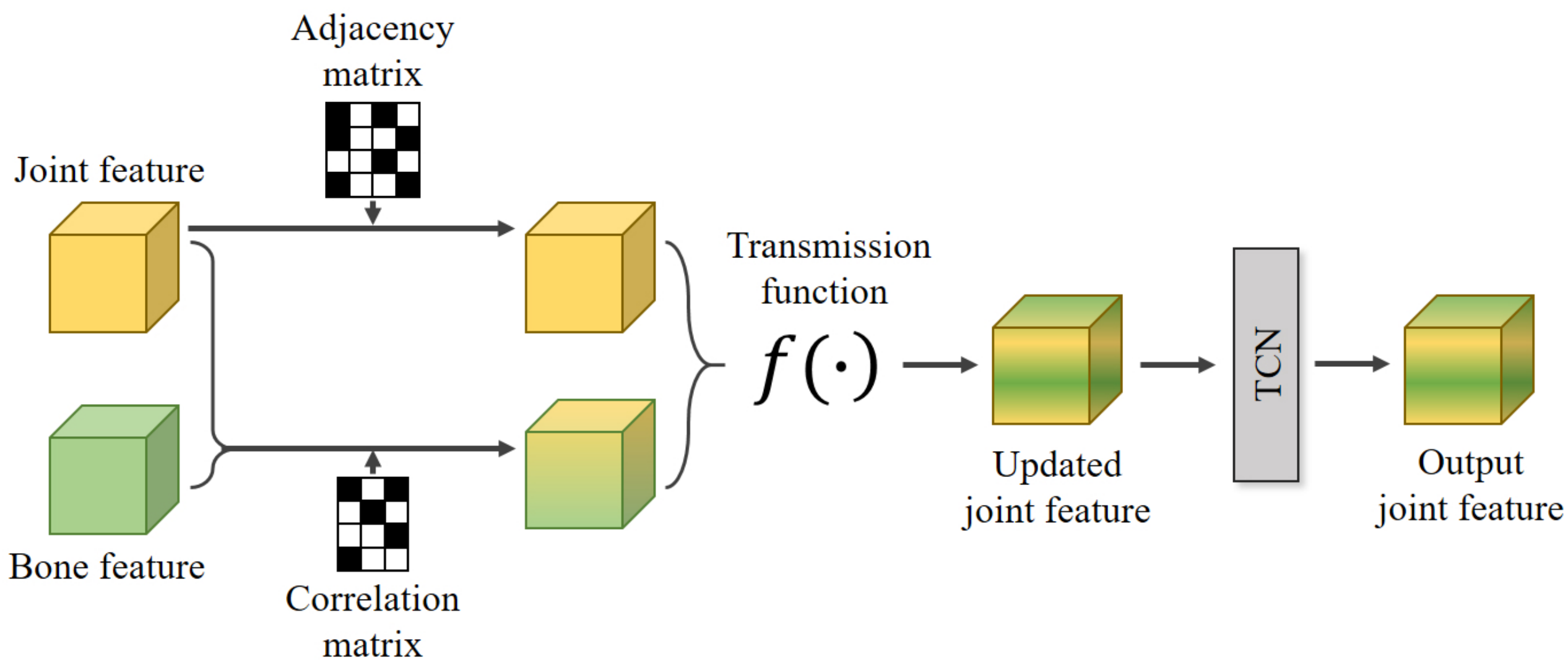

Fig. 4. The flow chart of the proposed CD-JBF-GC module in the joint stream. The CD-JBF-GC module can fuse the motion information of the bones into the joint stream with the help of the correlation matrix and the feature transmission function.

There is a natural connecting structure between the joints and bones of the human body. We can utilize a correlation matrix $V \in \mathbb{R}^{n\times m}$ to describe the connection structure of $n$ joints and $m$ bones even though they belong to different graphs. If the $i$-th joint and the $j$-th bone are connected, $V_{i,j} = 1$, otherwise $V_{i,j} = 0$. For one root joint, we divide its connected bones into two sets: 1) bones pointing to the root joint (in-bone); 2) bones away from the root joint (out-bone). We define the direction of the bone according to the same rule as [32], i.e. a bone starts from the joint that is closed to the barycenter of the human body and ends at the joint that far away from the human body barycenter. To correspond to the adjacency matrix $A$ which has three subsets, we add a third set initialized as a zero matrix for $V$. In this way, the partition group set is denoted as $P = \{zero, in, out\}$ and $\sum_{p\in P} V^p = V$. For the bone stream, we adopt the same strategy as the joint stream, and use $V^T \in \mathbb{R}^{m\times n}$ as the correlation matrix, which is the transpose of $V$.

For motion information transmission, the correlation matrix $V$ determines the direction while the feature transmission function $f(\cdot)$ determines the way of feature fusion. To evaluate the performance and the effect of the feature transmission function, we tried many different implementations, including element-wise summing, average-pooling, max-pooling, G-GRU, and feature concatenation. After that, we use a trainable weight $W^p \in \mathbb{R}^{d_{in}\times d_{out}}$ to fully exploit the function of feature transmission. The experimental comparison will be discussed in section 5.3.

### B. Model Architecture

Our model consists of two streams, i.e., a joint stream and a bone stream. The joint stream takes human body joints as graph vertexes and bones as graph edges to construct the skeleton graph sequence, and the initial feature of the vertex is its 3D coordinate corresponding to the human body joint. The bone stream takes human bones as graph vertexes and joints as graph edges, and the initial feature of the bone is the coordinate of the target joint minus the coordinate of the source joint, that is, the spatial difference of the joints. We define the joint, which closes to the centre of gravity of the skeleton, as the source joint; and define the joint, which is far away from the centre of gravity, as the target joint. For example, given a bone with its source joint $v_1 = (x_1, y_1, z_1)$ and its target joint $v_2 = (x_2, y_2, z_2)$, the initial feature of the bone is calculated as $v_2 - v_1 = (x_2 - x_1, y_2 - y_1, z_2 - z_1)$. The overall architecture of our proposed CD-JBF-GCN is shown in Fig. 5.

***Encoder***. The encoder, which consists of 10 layers (i.e. 5 ST-GC layers and 5 CD-JBF-GC layers), aims to extract rich spatial-temporal motion information from the skeleton sequence. For the skeleton sequence of $T$ frames, we use $\sigma T$ frames as the input of the encoder, where $0 < \sigma < 1$ is a hyper-parameter. Generally, the value of $\sigma$ should be larger than 0.5 to ensure the input sequence contains more complete action information. The dimension of the input feature is $\mathbb{R}^{n(m)\times 3\times \sigma T}$. After being processed by the encoder, the output feature dimension changes to $\mathbb{R}^{n(m)\times 256\times \frac{\sigma T}{4}}$, which is used as the input of the decoder and the classifier.

***Classifier***. The role of the classifier is to classify the human action of the skeleton sequence based on the motion characteristics extracted by the encoder. The classifier consists of a Spatial-temporal Average Pooling layer (ST-AVG-POOLING) and a Fully Connected layer (FC). Then, the output score of the joint stream and the bone stream are summed up as the final action classification score.

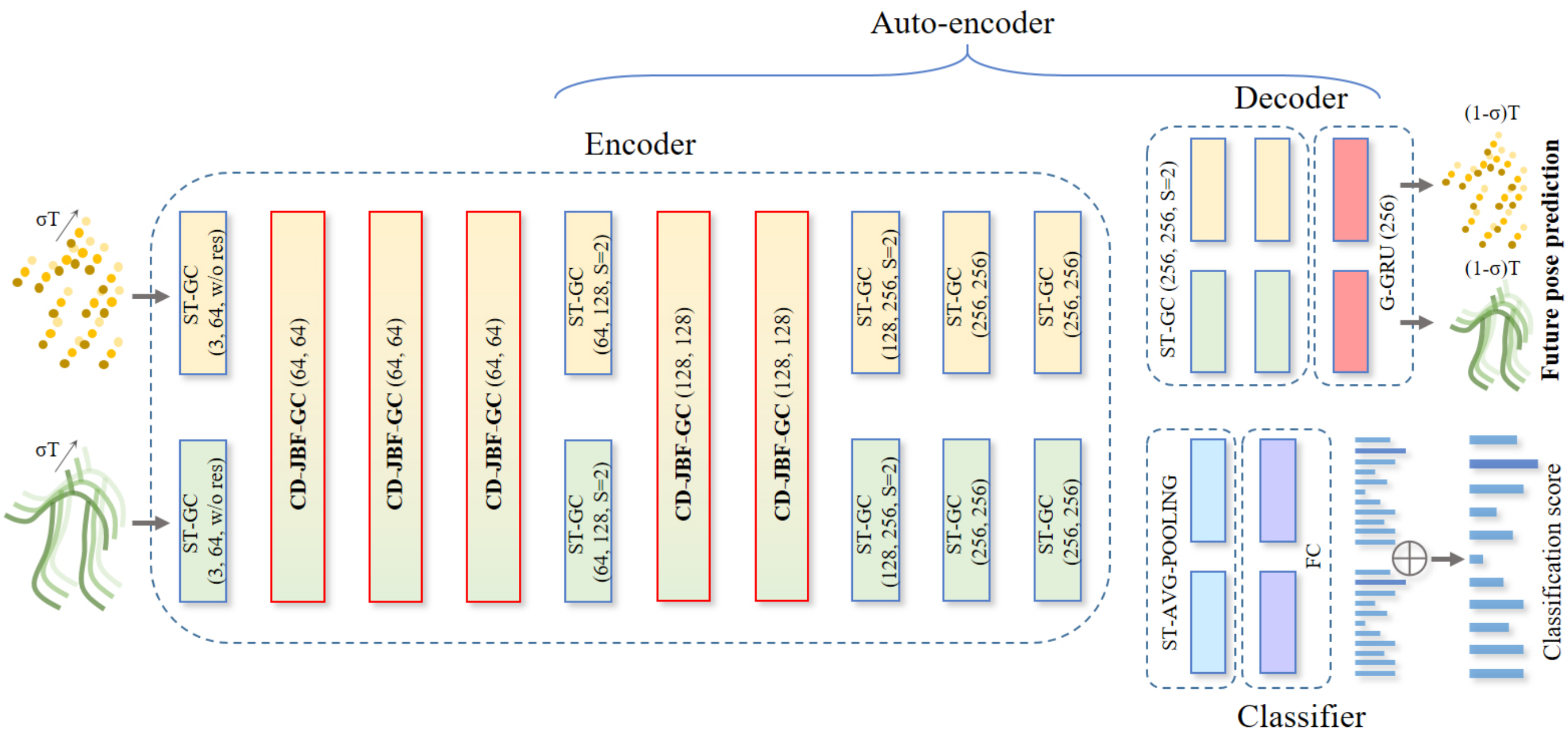

Fig. 5. The overall architecture of the proposed CD-JBF-GCN. The layer with annotation (3, 64, w/o res) means the input channel of this layer is 3, the output channel of this layer is 64, and this layer doesn't have a residual connection. "S=2" means the stride size of the layer is 2. Refer Fig. 2 to see the semi-supervised training strategy of our CD-JBF-GCN for detail.

***Decoder***. The role of the decoder is to predict the future spatial-temporal changes of the human joints and bones based on the features extracted by the encoder. The encoder and the proposed future pose prediction decoder can be trained by the self-supervised learning method to learn motion information from unlabeled data. The core module of the decoder is a Graph Gated Recurrent Unit (G-GRU). We use the same strategy as [58] to predict future $(1-\sigma)T$ frames of the joint and bone sequences, where the G-GRU takes two inputs: 1) the initial state $X^t \in \mathbb{R}^{n(m)\times 9}$ that includes the joint coordinates (bone vectors) at time $t$, its first-order and second-order difference, and 2) the online skeleton motion feature $H^t \in \mathbb{R}^{n(m)\times 256}$. Before the G-GRU, we add two ST-GC layers to bridge the gap between the input feature of the decoder and classifier.

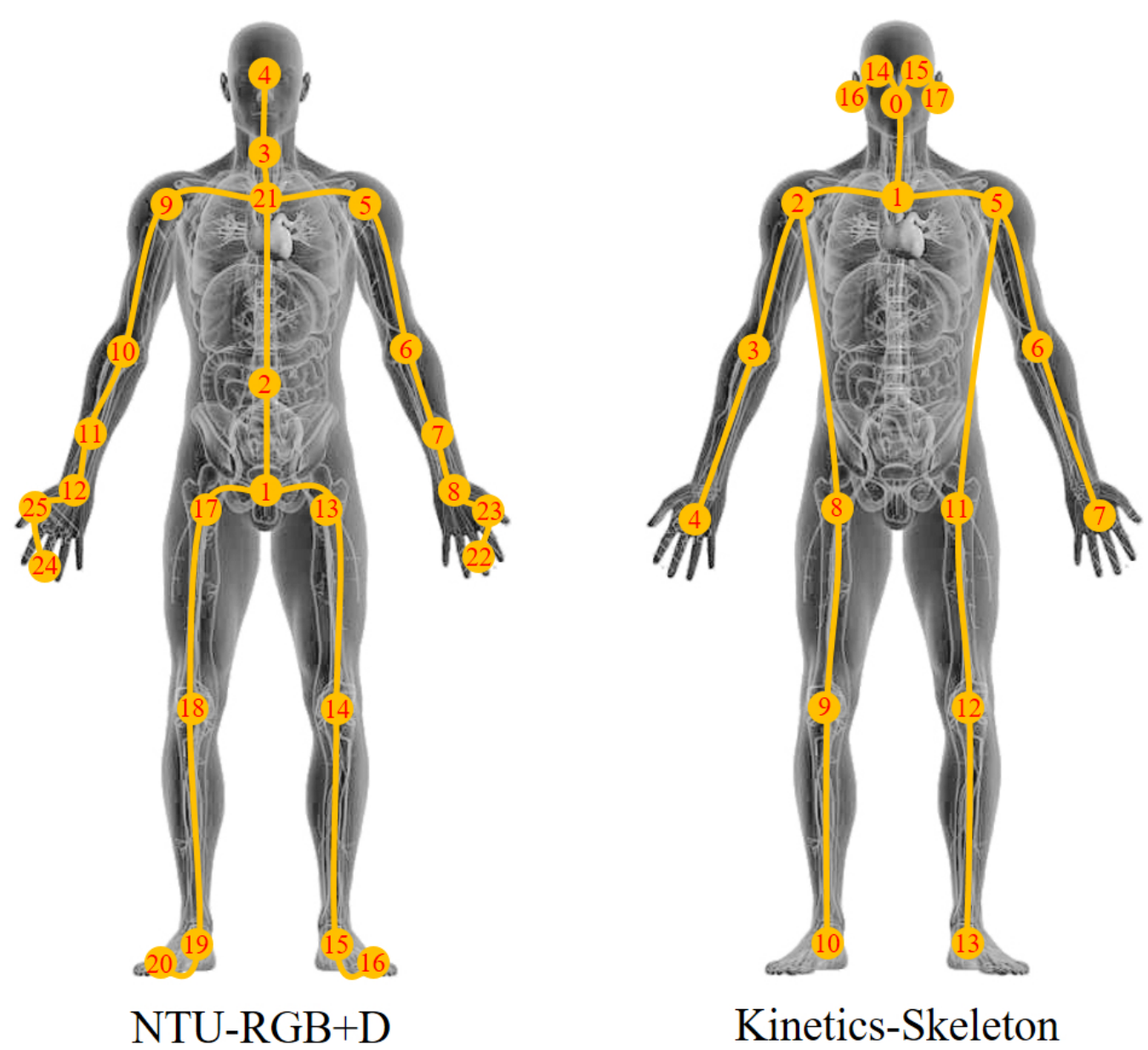

Fig. 6. Illustration of the human skeleton graphs and physical connections in the NTU-RGB+D dataset and the Kinetics-Skeleton dataset.

### C. Semi-supervised Training

Annotating ground-truth for skeleton-based action recognition is time-consuming. We introduce a semi-supervised training method to use less annotated data to train an applicable skeleton-based action recognition model by making full use of the unannotated data (see Fig. 2). When using the unannotated set $U$ to train the model, the decoder predicts future $(1-\sigma)T$ frames of the skeleton sequence, and this prediction is compared with the original skeleton sequence to compute an L1 loss $Loss_p$. In this case, the training loss $Loss_U$ is defined as:

$$Loss_U = Loss_p. \tag{6}$$

The L1 loss $Loss_p$ can be calculated as:

$$Loss_p = \frac{\sum_{i=1}^{N}\sum_{j=1}^{3}\sum_{k=1}^{T}|x^p_{i,j,k} - x_{i,j,k}|}{3TN}, \tag{7}$$

where $N$ is the number of the joints (bones) and $T$ is the sequence length. The unannotated training Loss $Loss_U$ is applied to optimize the encoder and the decoder. When using the annotated set $L$ to train the model, except the L1 loss $Loss_p$ which is computed by the decoder, the classifier is conducted to perform action classification and a Cross-entropy Loss $Loss_c$ is calculated. Accordingly, for the annotated data, the training loss $Loss_L$ is defined as:

$$Loss_L = Loss_c + \lambda Loss_p, \tag{8}$$

where $\lambda$ is a weighting factor. The training loss $Loss_L$ is used to optimize the whole network, including the encoder, decoder and classifier. The Cross-entropy Loss $Loss_c$ can be calculated as:

$$Loss_c = \sum_{c=1}^{M} p_c \log(q_c), \tag{9}$$

where $M$ is the number of the action classes, $p_c$ is the one-hot label, and $q_c$ is the classification score output by the model.

## V. Experiments

### A. Datasets

Two popular benchmark skeleton datasets, i. e. NTU RGB+D [62] and Kinetics-Skeleton [24], are selected for experimenting. The human skeleton graphs and physical connections in the NTU-RGB+D dataset and the Kinetics-Skeleton dataset can be seen in Fig. 6.

**NTU-RGB+D**: NTU-RGB+D is a large-scale dataset with annotated 3D joint coordinates of human body for the task of human action recognition [62]. NTU-RGB+D includes 56,000 action videos with the action categories of 60. These videos are in-door-captured from 40 volunteers in different age groups ranging from 10 to 35. For each action, the videos are captured by 3 cameras from different viewpoints, and the 3D annotations of human body joints are given in the camera coordinate system. Every action video has no more than 2 subjects and there are 25 key joints for each subject in the skeleton sequences.

The NTU-RGB+D dataset has two sub-sets: (1) Cross-Subject (CS) sub-set, which consists of 40,320 training videos and 16,560 testing videos. This sub-set contains 20 subjects, where 1 subject is utilized for training and the remaining 19 subjects are used for evaluation; (2) Cross-View (CV) sub-set, which composes of 37,920 training videos and 18,960 testing videos. In this sub-set, the video samples which come from camera viewpoints 2 and 3 are used for training, and the video samples, which are captured by the camera viewpoint 1, are utilized for evaluation. We follow the conventional setting of [62] and report the top-1 accuracy on both sub-sets.

**Kinetics-Skeleton**: Kinetics [63] includes 300,000 video clips with 400 action classes. The video clips in Kinetics, which are sourced from YouTube, are abundant and multifarious. The samples in this dataset are raw videos that are without the skeleton annotation. Yan *et al*. [24] used the OpenPose toolbox [16] to estimate the locations of 18 joints on every video frame and accordingly released the Kinetics-Skeleton dataset. In this dataset, all videos are firstly converted to a frame rate of 30fps with the resolution of $340 \times 256$. Then these re-scaled videos are employed to produce the 2D coordinates and the confidence score for 18 joints of each human body in terms of the OpenPose toolbox. Specially, for the multi-person video clips, two major individuals are chosen by calculating the average joint confidence. Based on the generated 2D coordinate and confidence score, a joint is represented as a three-element feature vector. For experimental evaluation, we report both the top-1 and top-5 accuracies on the testing set as Yan *et al*. [24].

TABLE I
Comparison of the top-1 and top-5 accuracy on the NTU-RGB+D X-V benchmark with different feature transmission functions.

| Transmission functions | Top-1 (%) | Top-5 (%) |
|---|---|---|
| Baseline | 91.42 | 98.79 |
| Element-wise summing | 91.21 | 98.68 |
| Average-pooling | 91.47 | 98.81 |
| Max-pooling | 91.55 | 98.81 |
| G-GRU | 91.90 | 98.87 |
| Feature concatenation | **92.43** | **99.08** |

### B. Implementation Details

We implement our model based on the PyTorch deep learning framework [64]. We apply the stochastic gradient descent (SGD) algorithm with Nesterov momentum (0.9) as the optimizer. The weight decay is set to 0.0005. We use 4 Nvidia GTX 1080Ti GPUs for the model training, and set the batch size to 24 and the weighting factor to $\lambda = 0.1$. For the NTU-RGB+D dataset, the learning rate is set as 0.1 and the number of training epoch is set as 60. The learning rate decay operates at the 30th epoch, 40th epoch and 50th epoch in training process and the factor is set as 0.1. For the Kinetics-Skeleton dataset, the number of training epoch is set to 70 and the learning rate is set to 0.1 as [31]. The learning rate decay is set to 0.1 at the 40th epoch, 50th epoch and 60th epoch.

### C. Self-comparisons

We present an ablative analysis on the NTU-RGB+D X-V benchmark to evaluate the effectiveness of the explored modules in our model. We randomly select 50% of the training data when to fully-supervised train the model, and use 100% of the testing data at the testing phase. We analyze the effect of the CD-JBF-GC, different encoder configurations, and the semi-supervised learning method.

**Joint-bone feature transmission**. We conduct experiments to test the performance of the feature transmission function. Several alternative functions are investigated including element-wise summing, average-pooling, max-pooling, G-GRU, and feature concatenation with the trainable weight $W$. When no transmission function is used, the joint stream and the bone stream are trained separately (We set it as the "baseline"). The results in Table I show that compared to the baseline, CD-JBF-GC is beneficial for action classification but not all feature transmission functions are suitable for expressing the correlation between joints and bones. The element-wise summing, average-pooling, and max-pooling almost have no effect on boosting the expressive ability of the model, the top-1 accuracy improvement is less than 0.2%. G-GRU and feature concatenation are all beneficial for joint-bone motion feature transmission and can obtain higher accuracy than the baseline. We think that G-GRU and feature concatenation can provide more learnable parameters to enable the model to better explore the potential functional correlation between joints and bones, and show stronger robustness to different human actions.

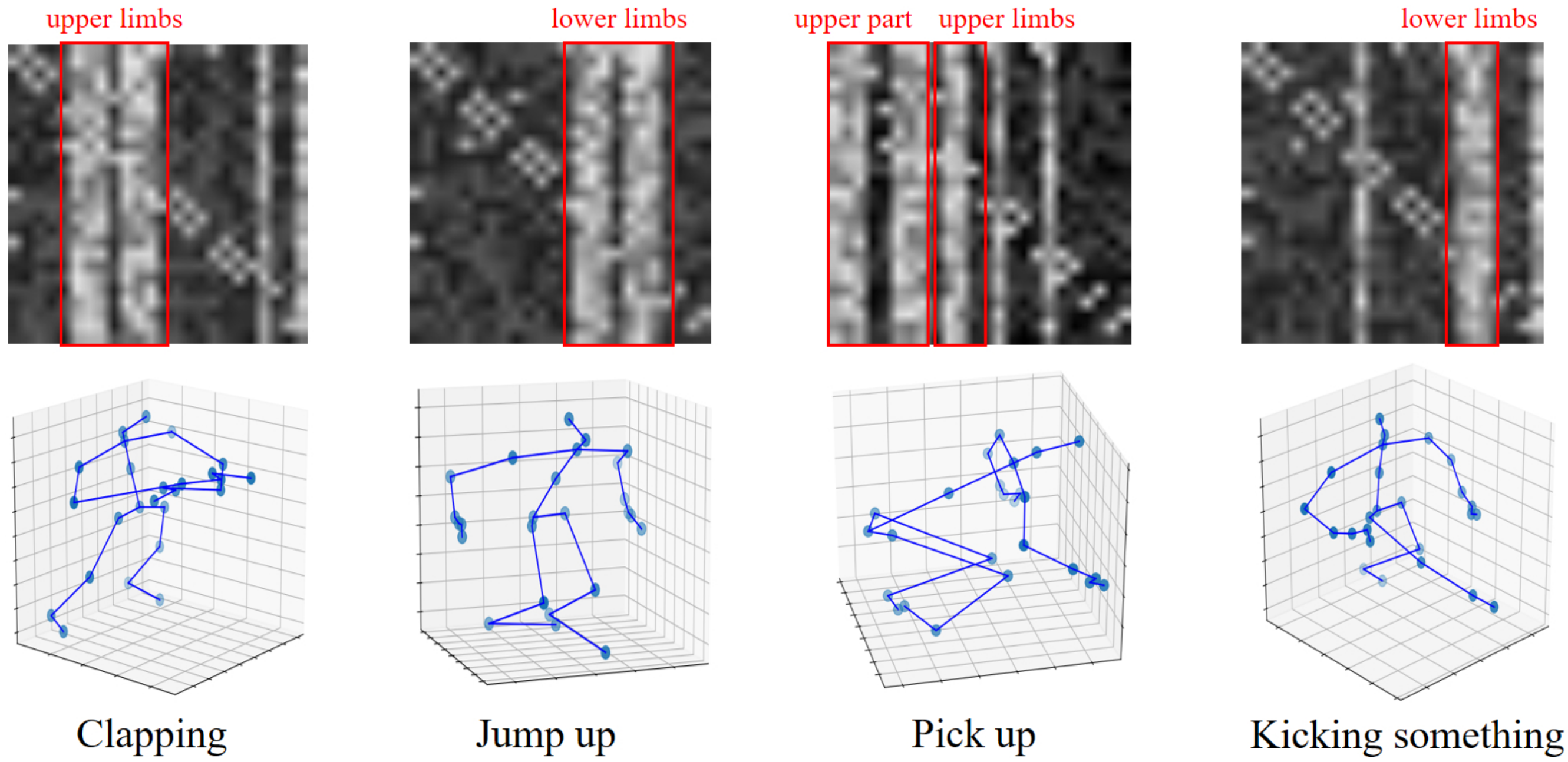

Fig. 7. Visualization of the correlation matrix in the joint stream CD-JBF-GCN for different human action classes on the NTU-RGB+D dataset. The value of the correlation matrix is the average of all the layers' correlation matrices in the model. The brighter area indicates that the weight of the correlation matrix is larger there, which means the correlation between the joint and the bone is stronger.

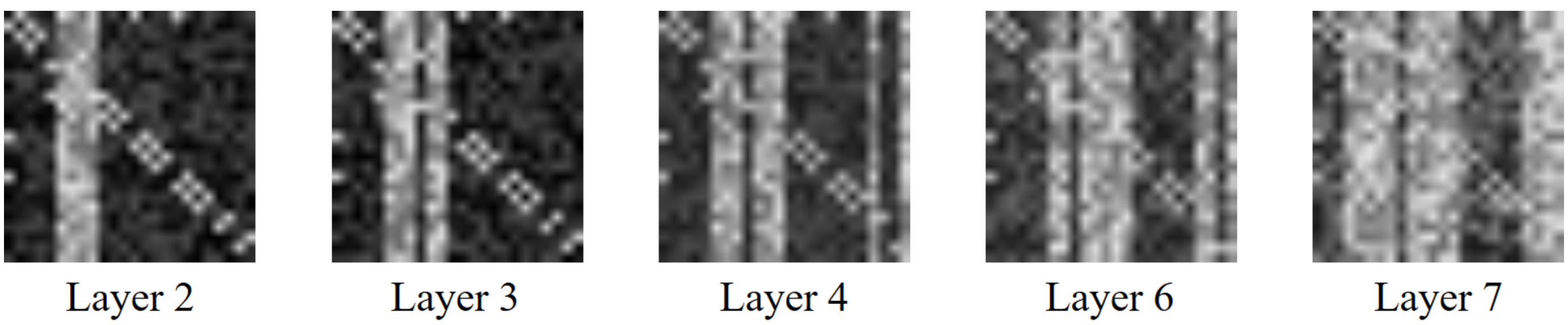

Fig. 8. Visualization of the correlation matrix in the 2nd, 3rd, 4th, 6th, and 7th layers of the joint stream CD-JBF-GCN for the human action "clapping hands". In the low-level correlation matrix, there is a strong correlation between the joints and the local bones which have significant motion information. In the high-level correlation matrix, the correlation between the joints and the bones is gradually smoothed.

As shown in Table I, among all transmission functions we tested, the feature concatenation has the best performance, where the top-1 accuracy is improved by 1.01% compared to the baseline. This is because after the feature concatenation, the weight $W$ can map the dimension of the concatenated feature to the previous scale, accordingly, the joint stream (bone stream) graph vertexes can receive more useful information from the bone stream (joint stream) and avoid the influence of redundant information. Based on the experimental results and the reasons given above, in our model, we choose the feature concatenation with trainable weight $W$ as the feature transmission function $f(\cdot)$ (see Eq. 5). Specifically, we use the following formula to update the features:

$$X_j^{'} = \sum_{p \in P} concat\left(A^p X_j, V^p X_b\right) W^p. \tag{10}$$

For motion information transmission, the feature transmission function $f(\cdot)$ determines the way of feature fusion, and the correlation matrix $V$ determines the direction of the motion transmission among the joints and bones. Therefore, whether the correlation matrix can learn useful joint-bone correlation is very important for the transmission of motion information. Fig. 7 shows the correlation matrix in the joint stream of CD-GBF-GCN for different action classes. For the NTU-RGB+D dataset, the dimension of the joint stream correlation matrix is $25 \times 24$. In Fig. 7, the value of the correlation matrix is the average of all the layers' correlation matrices in the model. The brighter area indicates that the weight of the correlation matrix is greater there, that is, the correlation between the joint and the bone is stronger. We can clearly see that for different actions, the correlation matrix can learn the obvious movement parts of human bones and guide the motion information transmit to the joints. Taking "clapping hands" as an example, the activation value of the bones of the upper limb is significantly higher than that of other regions,

TABLE II

COMPARISON OF THE TOP-1 AND TOP-5 ACCURACY ON THE NTU-RGB+D X-V BENCHMARK WITH DIFFERENT ENCODER CONFIGURATIONS. THE SECOND COLUMN 'REPLACED LAYERS' INDICATES THE NUMBER OF THE ST-GC LAYER REPLACED BY THE CD-JBF-GC LAYER IN THE BASELINE ENCODER.

| Encoder configurations | Replaced layers | Top-1 (%) | Top-5 (%) |
|---|---|---|---|
| Baseline, 10 ST-GC | - | 91.42 | 98.79 |
| 3 CD-JBF-GC + 7 ST-GC | 2,3,4 | 91.91 | 98.91 |
| 5 CD-JBF-GC + 5 ST-GC | 2,3,4,6,7 | **92.43** | **99.08** |
| 7 CD-JBF-GC + 3 ST-GC | 2,3,4,6,7,9,10 | 90.80 | 98.32 |

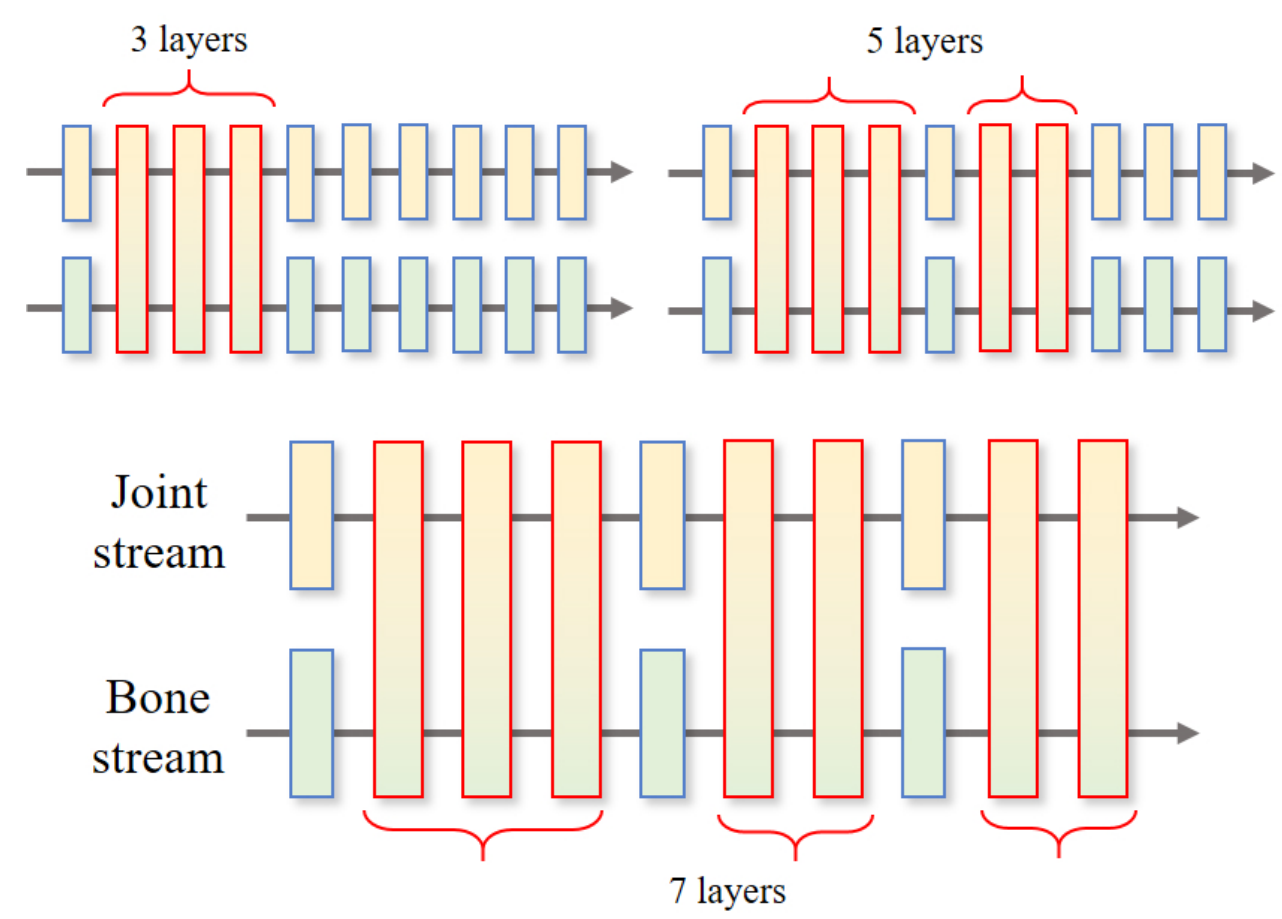

Fig. 9. Illustration of different encoder configurations. We gradually replaced 3, 5, and 7 ST-GC layers in the baseline with the CD-JBF-GC layers respectively, and tested the performance (see Table II).

which means that the bone movement of the upper limb is aggregated into the joint features, which effectively enhances the motion features of the joints. From another perspective, the effect of the correlation matrix is similar to a kind of attention map, which can make the model focuses on the distinct movement part of the human body and fully extracts the motion information with the help of the feature transmission function.

Fig. 8 shows the correlation matrix in different layers of the joint stream CD-GBF-GCN for “clapping hands”. It shows that when fusing low-level features, the correlation matrix focuses on the body parts with obvious motion information, but when fusing high-level features, the correlation matrix gradually becomes smooth. This because that due to the characteristics of GCN, when extracting high-level features, the features of the joints are gradually smoothed, so the activation area in the correlation matrix is not distinct. It also means that joints and bones will be more closely correlated in low-level features and our CD-JBF-GC layer is suitable for extracting low-level features of skeleton sequences.

**Encoder configurations**. Experiments are performed to test the impact of different encoder configurations. Taking the two-stream network with 10 ST-GC layers as the baseline. Fig. 8 shows that joints and bones are more closely correlated in low-level features, so we gradually replaced the ST-GC layers in the baseline with the CD-JBF-GC layers from low-level to high-level. The detailed encoder configurations can be seen in Fig. 9. Table II shows the results of replacing 3, 5, and 7 ST-GC layers with the CD-JBF-GC layers respectively. The configuration with 5 CD-JBF-GC layers and 5 ST-GC layers (see Fig. 5) achieves the best performance (respectively 1.01% and 0.29% higher than the baseline on top-1 and top-5 accuracy). The reason is that by using this structure, when extracting low-level features, the network can transmit and fuse joint-bone correlation information via CD-JBF-GC layers effectively. When extracting high-level features, the network can concentrate on processing joints and bones independently, which avoids excessive smoothing of the high-level features. Therefore, this model structure can not only fully exploit the respective motion information of joints and bones, but also can extract their mutual dependence. On the other side, when we use too many transmission layers, the performance of the model will decrease instead. For example, see the last row of Table II, where the results are respectively 0.62% and 0.47% lower than the baseline on top-1 and top-5 accuracy. This is because high-level skeleton motion features contain few correlations between joints and bones. Using a transmission layer to process high-level features will affect the independent feature expression of joints and bones.

**Semi-supervised learning**. We study the impact of semi-supervised learning of our method on the NTU-RGB+D X-V dataset and report top-1 accuracy. With the same network architecture, we use part of the annotated data from the training set to train the whole network and use the entire training data to train the auto-encoder without using their action labels. As shown in the 2-th column and the 4-th column of Table III, the performance is significantly promoted compared with the supervised learning (**SL**) method that is trained with the same size of annotated data as the semi-supervised learning (**SSL P**) methods. For example, with the usage of 1% labeled data, the semi-supervised learning (**SSL P**) method outperforms the supervised learning (**SL**) method by 9.56% (29.69% vs 20.13% ).

Furthermore, we apply the state-of-the-art skeleton inpainting head (I) [33], [34] as a decoder to replace the future pose prediction head (P) proposed in this work and compare their performance. Results in the 3-th column and 4-th column of Table III show that for GCN-based methods, the I head is unsuitable for semi-supervised learning because the missing key vertexes will affect the ability of GCN to extract motion features, which resulting in GCN cannot learn useful motion information. On the contrary, the P head is more suitable for GCN-based models to learn skeleton motion information in the semi-supervised way. Because our method avoids missing

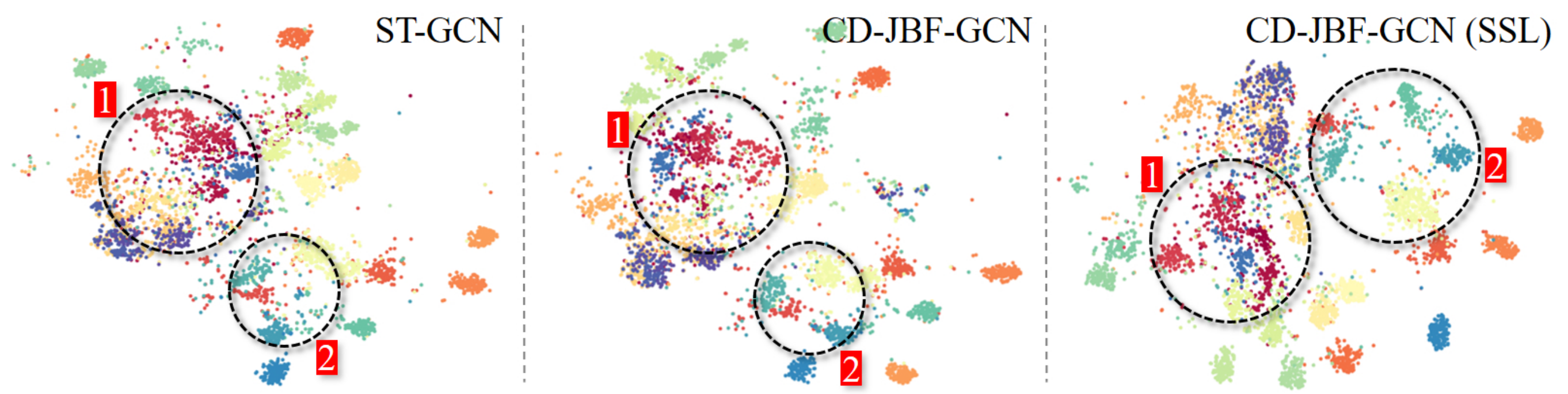

Fig. 10. t-SNE embeddings of the latent representation across ST-GCN and CD-JBF-GCN with 10% labeled data. Colors correspond to action classes. Two sample sets are denoted by dashed circles to show that our method can learn more discriminative category clusters.

TABLE III
COMPARISON OF THE TOP-1 ACCURACY ON THE NTU-RGB+D X-V BENCHMARK WITH DIFFERENT DECODER HEADS. "PERCENTAGE" MEANS THE PERCENTAGE OF THE LABELED TRAINING DATA USED. "SL" MEANS SUPERVISED LEARNING. "SSL I" MEANS SEMI-SUPERVISED LEARNING WITH SKELETON INPAINTING HEAD. "SSL P" MEANS SEMI-SUPERVISED LEARNING WITH FUTURE POSE PREDICTION HEAD.

| Percentage | SL (%) | SSL I (%) | SSL P (ours,%) |
|---|---|---|---|
| 50% | 92.09 | 91.54 | **92.43** |
| 10% | 76.50 | 75.91 | **78.03** |
| 5% | 61.69 | 61.06 | **65.28** |
| 1% | 20.13 | 19.72 | **29.69** |

TABLE IV
COMPARISON OF THE TOP-1 AND TOP-5 ACCURACY ON THE NTU-RGB+D X-V BENCHMARK WITH DIFFERENT VALUES OF $\sigma$.

| The value of $\sigma$ | Top-1 (%) | Top-5 (%) |
|---|---|---|
| 0.97 | 92.28 | 98.95 |
| 0.95 | **92.43** | **98.98** |
| 0.93 | 92.18 | 98.91 |

the key vertexes in the graph and will not affect the GCN to extract spatial-temporal feature.

For the hyper-parameter $\sigma$, we set it according to [58]. The experimental results in Table IV show that when $\sigma$ is around 0.95, its impact on the final performance becomes stable.

Fig. 10 shows the t-SNE embeddings of the latent representation across ST-GCN and CD-JBF-GCN with 10% labeled data. Two sample sets are denoted by dashed circles to show that our CD-JBF-GCN and the semi-supervised strategy can get more compact feature clusters and more obvious cluster spacings compared with ST-GCN. This intuitively verifies that our method can explore the motion correlation between joints and bones, and make full use of the unlabeled data to learn more discriminative features.

### D. Comparison with State-of-the-arts

We compare the proposed CD-JBF-GCN with the state-of-the-art skeleton-based action recognition methods on both the NTU-RGB+D and Kinetics-Skeleton datasets. The methods which are selected for comparison include the fully-supervised methods [3], [7], [24], [26], [28], [29], [32], [39], [40], [62], [65], [66] and the semi-supervised methods [33], [34], [50], [67]. Specifically, the fully-supervised methods include RNN-based algorithms [3], [7], [62], CNN-based algorithms [39], [40], [65], and GCN-based algorithms [24], [26], [28], [29], [32], [43], [66]. When comparing with the fully-supervised methods, we use all of the labeled training data. When comparing with the semi-supervised methods, we use the whole training data to train the auto-encoder without using their labels, then utilize part of the annotated data from the training set to train the entire network (including the encoder, decoder, and classifier).

The results of comparison with fully-supervised methods on the NTU-RGB+D dataset and the Kinetics-Skeleton dataset are shown in Table V and Table VI respectively. Notably, our CD-JBF-GCN performs better than ST-GCN [24] and 2s-AGCN [32], which two are the most relevant methods to us recently. For 2s-AGCN, our results outperform it by 0.8% (89.0% vs 88.2%) on the NTU-RGB+D X-S benchmark, 0.5% (95.4% vs 94.9%) on the NTU-RGB+D X-V benchmark, and 0.6% (36.5% vs 35.9%) on the Kinetics-Skeleton dataset, respectively. This reveals that our CD-JBF-GCN model can better classify a variety of human actions by using the CD-JBF-GC module.

TABLE V
COMPARISON OF THE TOP-1 ACCURACY WITH STATE-OF-THE-ART FULLY-SUPERVISED METHODS ON THE NTU-RGB+D DATASET. "X-S" MEANS CROSS-SUBJECT, "X-V" MEANS CROSS-VIEW.

| Methods | X-S (%) | X-V (%) |
|---|---|---|
| HBRNN (2015) [7] | 59.1 | 64.0 |
| Deep LSTM (2016) [62] | 60.7 | 67.3 |
| ST LSTM (2016) [3] | 67.2 | 77.7 |
| TCN (2017) [65] | 74.3 | 83.1 |
| Synthesized CNN (2017) [40] | 80.0 | 87.2 |
| CNN+M+T (2017) [39] | 83.2 | 89.3 |
| ST-GCN (2018) [24] | 81.5 | 88.3 |
| M-GCNs+VTDB (2019) [66] | 84.2 | 94.2 |
| AS-GCN (2019) [26] | 86.8 | 94.2 |
| 2s-AGCN (2019) [32] | 88.2 | 94.9 |
| CA-GCN (2020) [29] | 83.5 | 91.4 |
| SGN (2020) [28] | 89.0 | 94.5 |
| 2s-Shift-GCN (2020) [43] | 89.7 | 96.0 |
| CD-JBF-GCN (Ours) | **89.0** | **95.4** |

TABLE VI
COMPARISON OF THE TOP-1 AND TOP-5 ACCURACY WITH STATE-OF-THE-ART FULLY-SUPERVISED METHODS ON THE KINETICS-SKELETON DATASET.

| Methods | Top-1 (%) | Top-5 (%) |
|---|---|---|
| Deep LSTM (2016) [62] | 16.4 | 35.3 |
| ST-GCN (2018) [24] | 30.7 | 52.8 |
| AS-GCN (2019) [26] | 34.8 | 56.5 |
| 2s-AGCN (2019) [32] | 35.9 | 58.6 |
| CA-GCN (2020) [29] | 34.1 | 56.6 |
| CD-JBF-GCN (Ours) | **36.5** | **59.6** |

TABLE VII
COMPARISON OF THE TOP-1 ACCURACY WITH STATE-OF-THE-ART SEMI-SUPERVISED METHODS ON THE NTU-RGB+D X-V AND X-S BENCHMARK.

| Methods | 5% | | 10% | | 20% | | 40% | |
|---|---|---|---|---|---|---|---|---|
| | X-V | X-S | X-V | X-S | X-V | X-S | X-V | X-S |
| VAT (2018) [67] | 57.9 | 51.3 | 66.3 | 60.3 | 72.6 | 65.6 | 78.6 | 70.4 |
| $S^4L$ (2019) [34] | 55.1 | 48.4 | 63.6 | 58.1 | 71.1 | 63.1 | 76.9 | 68.2 |
| ASSL (2020) [33] | 63.6 | 57.3 | 69.8 | 64.3 | 74.7 | 68.0 | 80.0 | 72.3 |
| $MS^2L$ (2020) [50] | - | - | - | 65.2 | - | - | - | - |
| CD-JBF-GCN (Ours) | **65.3** | **61.8** | **78.0** | **71.7** | **85.9** | **78.4** | **90.9** | **83.2** |

The results of comparison with semi-supervised methods on the NTU-RGB+D X-V benchmark and the X-S benchmark are shown in Table VII. Our CD-JBF-GCN with future pose prediction head outperforms the previous semi-supervised methods by a significant margin. Specifically, on the X-V benchmark, compared to the state-of-the-art method ASSL [33], our CD-JBF-GCN surpasses it by 1.7%, 8.2%, 11.2%, and 10.9% with using 5%, 10%, 20%, and 40% of the labeled training data respectively. On the X-S benchmark, compared to ASSL, the improvements reach to 4.5%, 7.4%, 10.4%, and 10.9% with using 5%, 10%, 20%, and 40% of the labeled training data respectively. Compared to the multi-task semi-supervised approach $MS^2L$ [50], our method pays more attention to the action recognition task and surpasses it by 6.5% on the X-S benchmark. The experimental results demonstrate that our CD-JBF-GCN model with future pose prediction head can effectively learn rich motion representations of human actions from the unlabeled data.

## VI. CONCLUSIONS

In this work, we proposed a novel correlation-driven joint-bone fusion graph convolutional network (CD-JBF-GCN) and designed a pose prediction head in self-supervised learning for skeleton-based human action recognition. The CD-JBF-GC module can explore the functional correlation between the joint stream and the bone stream, which is helpful to promote both streams to learn more discriminative feature representations. We tested and analyzed the CD-JBF-GC module, and found that the feature concatenation with a trainable weight is an appropriate transmission function which can significantly enhance the feature learning capability of the network. Moreover, we combine the self-supervised learning of pose prediction with the supervised learning of action classification, and explored a novel semi-supervised model to fully extract motion information from the skeleton data. Extensive experiments are conducted on two large-scale datasets to evaluate the performance of our method on recognizing human actions, results demonstrated that the proposed model exceeds the state-of-the-art semi-supervised methods by a large margin, which is also beneficial for fully-supervised learning.

## ACKNOWLEDGMENT

This work was supported by the National Natural Science Foundation of China under Grant 62106177. It was also supported by the Central University Basic Research Fund of China (No.2042020KF0016). The numerical calculation was supported by the supercomputing system in the Supercomputing Center of Wuhan University.

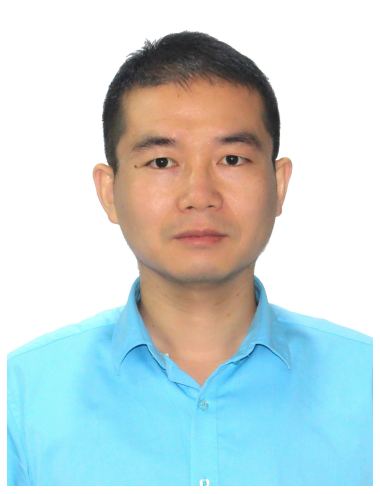

**Zhigang Tu** started his Master Degree in image processing at the School of Electronic Information, Wuhan University, China, 2008. In 2015, he received the Ph.D. degree in Computer Science from Utrecht University, Netherlands. From 2015 to 2016, he was a postdoctoral researcher at Arizona State University, US. Then from 2016 to 2018, he was a research fellow at the School of EEE, Nanyang Technological University, Singapore. He is currently a professor at the State Key Laboratory of Information Engineering in Surveying, Mapping and Remote sensing, Wuhan University. His research interests include computer vision, image processing, video analytics, and machine learning. Special for motion estimation, video super-resolution, object segmentation, action recognition and localization, and anomaly detection. He has co-/authored more than 50 articles on international SCI-indexed journals and conferences. He is an Associate Editor of the SCI-indexed journal TVC (IF=1.456), a Guest Editor of JVCIR (IF=2.479) and Combinatorial Chemistry and High Throughput Screening (IF=1.195). He is the first organizer of the ACCV2020 Workshop on MMHAU (Japan). He received the "Best Student Paper" Award in the $4^{th}$ Asian Conference on Artificial Intelligence Technology.

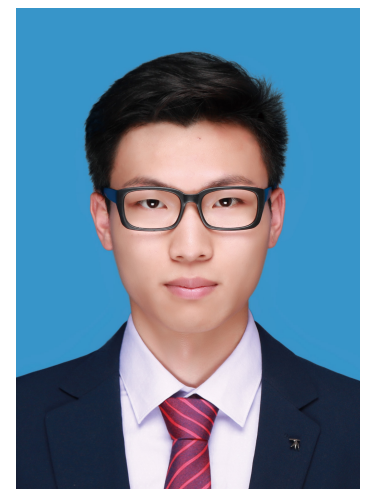

**Jiaxu Zhang** received the B.S. degree from Southeast University, Nanjing, China, in 2020. He is currently working toward the M.S. degree at the LIESMARS (State Key Laboratory of Information Engineering in Surveying, Mapping and Remote Sensing), Wuhan University, China. His research interests include computer vision, deep learning, and action recognition.

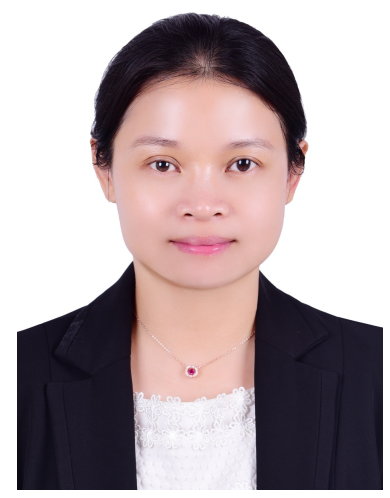

**Hongyan Li** received the master's degree in computer science from Central China Normal University, China, and the Ph.D. degree in computer architecture from the Huazhong University of Science and Technology, China, in 2016. Since 2017, she has been a Post-Doctoral Researcher with the State Key Laboratory of Information Engineering in Surveying, Mapping and Remote Sensing, Wuhan University. She is currently a Professor with the Hubei University of Economics. Her research interests include computer vision, big data, and machine learning.

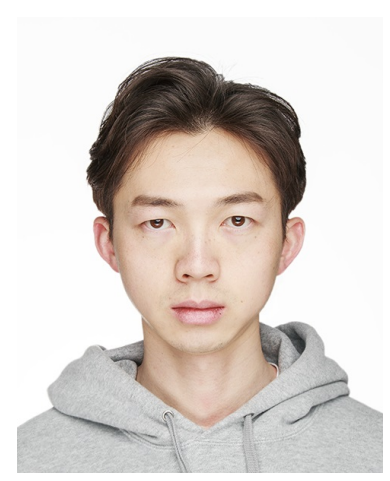

**Yujin Chen** received the B.Eng. and Ms. degree in geo-Information from Wuhan University, China, 2018 and 2021. He is currently a Ph.D. student at the Technical University of Munich, Germany. His research interests include computer vision and machine learning.

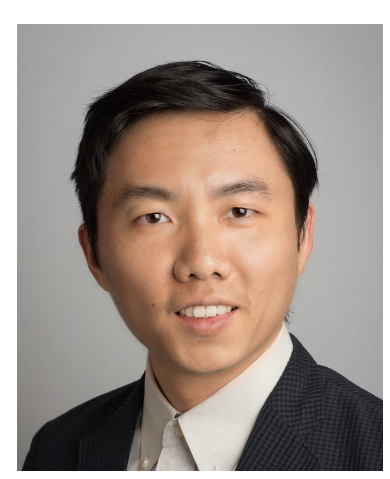

**Junsong Yuan** (M'08-SM'14) received his Ph.D. from Northwestern University and M.Eng. from National University of Singapore. He is currently an associate professor at Computer Science and Engineering department of State University of New York at Buffalo. Before that, he was an associate professor at Nanyang Technological University (NTU), Singapore. His research interests include computer vision, video analytics, gesture and action analysis. He received best paper award from Intl. Conf. on Advanced Robotics (ICAR'17), 2016 Best Paper Award from IEEE Trans. on Multimedia, Doctoral Spotlight Award from IEEE Conf. on Computer Vision and Pattern Recognition (CVPR'09), and outstanding EECS Ph.D. Thesis award from Northwestern University.

He is currently Senior Area Editor of Journal of Vis. Communications and Image Repres. (JVCI), Associate Editor of IEEE Trans. on Image Processing (T-IP) and IEEE Trans. on Circuits and Systems for Video Technology (T-CSVT). He is Program Co-chair of ICME'18, and Area Chair of CVPR'17/19/20/21, ACM MM'18, ICPR'18, ICIP'18'17, ACCV'18'14 etc. He is a Fellow of IEEE and International Association of Pattern Recognition (IAPR).